\documentclass{article}
\usepackage{ijcai25}

\usepackage{times}
\usepackage{soul}
\usepackage{url}
\usepackage[hidelinks]{hyperref}
\usepackage[utf8]{inputenc}
\usepackage[small]{caption}
\usepackage{graphicx}
\usepackage{amsmath}
\usepackage{amsthm}
\usepackage{booktabs}
\usepackage[switch]{lineno}

\usepackage{algorithm,algpseudocode}
\usepackage{algorithmicx}
\usepackage{mathtools}

\usepackage{latexsym}
\usepackage{amsfonts,amssymb}

\newtheorem{theorem}{Theorem}
\newtheorem{lemma}{Lemma}
\newtheorem{DEFINITION}[theorem]{Definition}
\newenvironment{definition}{\begin{DEFINITION} \rm }
	{\end{DEFINITION}}

\newcommand{\at}{ {\cal D} }
\newcommand{\calE}{ {\cal E}}
\newcommand{\emodels}{ \models_{\cal E}}
\newcommand{\KB}{\Gamma}
\newcommand{\NF}{ {\cal N\!F}}

\newcommand{\commentout}[1]{}

\newcommand{\imply}{\rightarrow}
\newcommand{\liff}{\leftrightarrow}

\newcommand{\eq}{\!=\!}

\newcommand{\lland}{\!\land\!}
\newcommand{\llor}{\!\lor\!}

\newcommand{\isdef}{\hbox{$\stackrel{{\scriptstyle def}}{=}$}}
\newcommand{\sa}{S_{\alpha}}
\newcommand{\bat}{\mathcal{D}}
\newcommand{\dsz}{\mathcal{D}_{S_0}}
\newcommand{\dsa}{\mathcal{D}_{\sa}}

\newcommand{\movebtob}{\mbox{\it move-b-to-b}}
\newcommand{\movebtot}{\mbox{\it move-b-to-t}}
\newcommand{\movettob}{\mbox{\it move-t-to-b}}

\mathchardef\mhyphen="2D

\renewcommand{\algorithmicrequire}{\textbf{Input: }}
\renewcommand{\algorithmicensure}{\textbf{Output: }}

\algnewcommand\algorithmicdefn{\textbf{Definitions:}}
\algnewcommand\Defn{\item[\algorithmicdefn]}

\title{\vspace*{-9mm} Planning with Dynamically Changing Domains}  

\author{
Mikhail Soutchanski$^1$
\and
Yongmei Liu$^2$
\affiliations
$^1$Toronto Metropolitan University (TMU), Toronto, Canada\\
$^2$Sun Yat-sen University, Guangzhou, China\\
}

\begin{document}

\maketitle

\begin{abstract}
In classical planning and conformant planning, it is assumed that there are finitely many named objects given in advance, and only they can participate in actions and in fluents. This is the Domain Closure Assumption (DCA). However, there are practical planning problems where the set of objects changes dynamically as actions are performed. We formulate the planning problem in first-order logic, assume an initial theory is a finite consistent set of fluent literals, discuss when this guarantees that in every situation there are only finitely many possible actions, impose a finite integer bound on the length of the plan, and propose to organize search over sequences of actions that are grounded at planning time.
We show the soundness and completeness of our approach. It can be used to solve the bounded planning problems without DCA that belong to the intersection of sequential generalized planning (without sensing actions) and conformant planning, restricted to the case without the disjunction over fluent literals. We discuss a proof-of-the-concept implementation of our planner.
\end{abstract}


\section{Introduction}
In a generalization of the British TV show Countdown (numbers round), the task is to produce the target number out of six given integers using only standard arithmetical operations with the condition that once a number has been used, it is no longer available for subsequent calculations. This is an instance of a planning problem in the case when fluents and actions have parameters that vary over infinite domains, knowledge about an initial state is incomplete, and objects can be created or destroyed during planning. This requires planning for dynamically changing domains. A planning algorithm has to search for a correct sequential plan with respect to several different infinitely sized logical models. It turns out that the foundations for this kind of planning research were developed years ago, as we explain in the next paragraph.

The real world around us can sometimes be conveniently understood in computational terms as a very large symbolic data base with incomplete knowledge, since a logical reconstruction of this open-world data base is a scalable logical theory that supports computationally tractable (sometimes) reasoning about incomplete knowledge. This perspective emerged in the early years of knowledge representation, for example, see   \cite{ReiterJACM1980,ReiterJACM1986,LinReiter1997,Levesque1998,LiuLevesqueIJCAI2003}, and inspired significant subsequent research, for example, in description logic \cite{DLHB2003}. 
Surprisingly, it has not been widely explored as much as should be in automated sequential planning with deterministic actions 
\cite{GhallabNauTraverso2004,GeffnerBonetBook2013}.

  There are conceptual connections between our research and several popular directions in modern planning. We clarify these connections next before explaining our contribution.

In classical and conformant planning, it is usually assumed that there are finitely many explicitly named objects given in advance, and that no objects are created or destroyed: this is the Domain Closure Assumption (DCA) \cite{ReiterJACM1980}. Moreover, in conformant planning, incomplete knowledge about an initial state is usually formulated using fluent disjunctions, but this requires combinatorial case analysis \cite{HoffmannBrafmanAIJ2006,PalaciosGeffnerJAIR2009,GrastienScalaAI2020}. These articles include the case where an initial state is a finite consistent set of fluent \textit{literals}, that is, there are fluents known to be true and fluents known to be false, but these and other papers implicitly require DCA. 
To the best of our knowledge, the case of planning with incomplete knowledge based on fluent literals without DCA has not been explored in conformant planning. 

Moreover, in classical, conformant and numerical planning, it is common to restrict the parameters of fluents and actions to finite domains, despite that there are practical examples of the planning problems that require actions with control parameters that vary over infinite domains (such as fuel, weight); for example, see  \cite{SavasFoxLongMagazzeniECAI2016}.

In the active research area of generalized planning (GP), the planning problem has to be solved for multiple instances at once.
To give a few examples,  research has ranged from contingent iterative planning (with sensing) 
\cite{HuLevesqueKR2010,SrivastavaPhD2010,BelleLevesqueKR2016}, 
to using abstractions \cite{BonetGeffnerJAIR2020,IllanesMcIlraith2019},
 exploring how generalized planning can be done with existentially quantified goals \cite{FunkquistStahlbergGeffnerKR2024},
or GP as a heuristic search approach that uses a GP program to solve a set of classical planning instances \cite{Segovia-AguasCelorrioJonssonAI2024}.
 The latter paper and \cite{CelorrioSegovia-AguasJonssonKER2019} review previous work in GP. 

We consider the generalized planning problem in first-order logic, with an (existentially quantified) goal formula, as a search for a plan from a consistent set of fluent literals and additional axioms, without the DCA, and compute a sequential plan with deterministic actions when it exists. This plan can be applied to any (infinitely-sized)  model of an initial incomplete logical theory. This planning problem requires computing action preconditions at run-time. In our setting, heuristic search over state space is infeasible since there are infinitely many models (states), and actions -- as well fluents -- may have parameters that vary over infinite domains. Therefore, we search for a plan over sequences of instantiated actions and formulate the conditions that guarantee, at every planning step, that there are only finitely many actions possible. If the length of a plan is bounded, then there are only finitely many sequences of  possible action that can be considered. We adapt a version of a domain-independent heuristic developed for the FF planner \cite{HoffmannNebel2001} to facilitate automated search.

Our approach can be used to solve the bounded planning problems (without DCA) that belong to the intersection of sequential generalized planning (with deterministic actions) and conformant planning, restricted to the case without the disjunction over fluent literals. We discuss a proof-of-the-concept implementation of our heuristic planner and demonstrate that it can solve a few instances without DCA.

\section{Preliminaries}
\subsection{Proper Initial Theory and the Procedure $V$}

We use a standard first-order (FO) language with equality, a countably infinite set of constants  $\{C_1,C_2,\ldots\}$, and no other function symbols. 
We restrict our attention to standard interpretations, where equality is identity, and there is a
bijection between the set of constants and the domain of discourse. This restriction can be captured by a set of axioms $\calE$, consisting of the axioms of equality and the set of formulas $\{C_i \neq C_j \mid i\neq j\}$.
We use $\rho$ to range over atoms whose arguments are distinct
variables. We use $e$ to range over equality well-formed formulas (ewffs), that is, quantifier-free formulas with only equalities (i.e. no predicates). We use $\forall \phi$ to denote the universal closure of $\phi$. We let $\theta$ range over substitutions of
all variables by constants, and write $\phi\theta$ as the result of applying $\theta$ to $\phi$. We use the letter $d$ to range over the constants. We
write $\KB \emodels \phi$ to denote $\calE \cup \KB \models \phi$.

Intuitively, a proper theory $\KB$ represents a consistent set of ground literals. It generalizes databases by allowing incomplete knowledge about some of the elements of the domain. For example, in an infinite domain of trees that have sizes, consider axioms 
$\forall x,t(t\!=\!C_{127} \land x\!=\!4 \rightarrow size(t,x,S_0))$
and $\forall x,t(t\!=\!C_{1009} \land x\!=\!16 \rightarrow \neg size(t,x,S_0))$, i.e.,
the size of the specific tree  $C_{127}$ is 4, the size of the tree $C_{1009}$ 
is not 16, and there are no axioms about any other trees or their sizes. 

\begin{definition}
	A theory $\KB $ is proper if $\calE \cup \KB$ is consistent and $\KB$ is a finite set of formulas of the form $\forall (e\imply \rho)$ or $\forall (e\imply \neg \rho)$. 
\end{definition}

Since the logical entailment problem
for proper theories is undecidable, Levesque \cite{Levesque1998} proposes a reasoning scheme called $V$ for them.
Given a proper theory and a query,
$V$ returns one of three values 0 (known false), 1 (known true), or
$\frac{1}{2}$ (unknown) as follows:
\begin{enumerate}
\item $V[\KB, \rho \theta]=
\left\{ \begin{array}{ll} 1 & \mbox{if there is a $\forall(e\imply 
\rho)$ in $\KB$ s.t.}\\ 
& \mbox{$V[\KB, e \theta]=1$}\\
0 & \mbox{if there is a $\forall(e\imply \neg \rho)$ in $\KB$
s.t. }\\ & \mbox{$V[\KB, e \theta]=1$}\\
\frac{1}{2}& \mbox{otherwise}
\end{array}
\right. $.\\
\vspace*{1mm}\item $V[\KB,(d=d')]=1$ if $d$ and $d'$ are identical constants,
and 0 otherwise; \vspace*{1mm}\item $V[\KB,\neg \phi]=1-V[\KB,\phi]$; \vspace*{1mm}\item
$V[\KB,(\phi \vee \psi)]\!=\!\max \{V[\KB,\phi], V[\KB,\psi]\}$; \vspace*{1mm}\item
$V[\KB,\exists x\phi]=\max\{V[\KB,\phi^x_d] \mid d\in H_1^+(\KB\cup\{\phi\})\}$,  
where $\phi^x_d$ means substitution of $d$ for $x$,
$H_1^+(\KB\cup\{\phi\})$ includes all constants from $\KB$, from  $\phi$,  and one extra constant  
that represents infinitely many objects not mentioned in 
$\KB\cup\{\phi\}$; they are indistinguishable. 
\end{enumerate}
Note that if a query includes $m$ quantifiers, then $m$ extra constants can be required to answer the query. 

$V$ is sound but incomplete. Levesque  proposes a certain normal form called $\NF$, and shows that $V$ is complete for queries in $\NF$: 
\begin{theorem} \cite{Levesque1998} Suppose $\KB$ is proper. Then for every $\phi$ in $\NF$, $\KB \emodels \phi$ iff $V[\KB,\phi]=1$; and $\KB \emodels \neg \phi$ iff $V[\KB,\phi]=0$.
\end{theorem}

We will not review the definition of $\NF$. The intuition behind $\NF$ is that different parts of a formula
must be logically independent; an example of a formula not in $\NF$ is  $(p\vee \neg p)$.

\subsection{The Situation Calculus}
The situation calculus (SC) is a logical approach to representation and reasoning about actions and their effects. It was introduced by John McCarthy,
\commentout{
}
and was refined by Reiter \cite{Reiter2001} who introduced the Basic Action Theories (BAT). 
Unlike the notion of state that is common in model-based planning, SC is based on the situation, namely a sequence of actions, 
which is a concise symbolic representation and a convenient proxy
for the state in the cases where all actions are deterministic
\cite{LevesquePirriReiter1998,LinKRhandbook2008}. 
We use variables $s,s',s_1,s_2$ for situations, variables $a,a'$ for actions,
and $\bar{x},\bar{y}$ for tuples of object variables. 
The constant $S_0$ represents the initial situation, and the successor function 
$do: action\times situation \mapsto situation$, e.g., $do(a,s)$, denotes 
situation that results from doing action $a$ in previous situation $s$. 
The terms $\sigma,\sigma'$ denote situation terms and
$A_i(\bar{x})$, or $\alpha,\alpha_1,\alpha_2,\alpha'$, represent
action functions and action terms, respectively. 
The shorthand $do([\alpha_1,\cdots,\alpha_n],S_0))$
represents the situation $do(\alpha_n,do(\cdots, do(\alpha_1,S_0)\cdots)$
resulting from the execution of actions $\alpha_1,\cdots,\alpha_n$ in $S_0$.
The relation $\sigma \sqsubset \sigma'$ between situations terms $\sigma$ and 
$\sigma'$ means that $\sigma$ is an initial subsequence of $\sigma'$. 
Any predicate symbol $F(\bar{x},s)$
with exactly one situation argument $s$ and possibly a tuple of object arguments
$\bar{x}$ is called a (relational) fluent. Without loss of generality, 
we consider only relational fluents.
A first-order logic formula $\psi(s)$
is called
\textit{uniform in $s$} if all fluents in $\psi$ mention only situation $s$ 
as their situation argument, and $\psi(s)$ has no quantifiers over the situation variables.

The basic action theory (BAT) ${\cal D}$ is the conjunction of the following classes of axioms 
\(
{\cal D}\!=\!\Sigma \land {\cal D}_{ss} \land {\cal D}_{ap} \land {\cal D}_{una}\land {\cal D}_{S_0}.\)
\\We use examples from the Blocks World (BW)  
domain \cite{LinReiter1997,LiuLakemeyer2009}. 
For brevity, all variables \ $\bar{x},a,s$ 
 are assumed $\forall$-quantified at front.

$\mathbf{{\cal D}_{ap}}$ is a set of action precondition axioms of the form\\
$\hspace*{0.5in} 
poss(A(\bar{x}), s)\liff \Pi_A(\bar{x},s),$\\
where $poss(a,s)$ is a new predicate symbol that means that an action $a$ is
possible in situation $s$,  $\Pi_A(\bar{x},s)$ is a formula uniform in $s$, 
and $A$ is an n-ary action function. 
In most planning benchmarks, the formula $\Pi_A$ is simply 
a conjunction of fluent literals and possibly negations of equality. 
We consider a version of BW, where there are three actions: 
$\movebtob(x,y,z)$, move a block $x$ from a block $y$ to another block $z$, 
$\movebtot(x,y)$, move a block $x$ from a block $y$ to the table, 
$\movettob(x,z)$, move a block $x$ from the table to $z$.
\\
$
\hspace{-2mm}
\begin{array}{l}
poss(\movebtob(x,\!y,\!z),s) \liff  clear(x,\!s)\land clear(z,\!s)\land\\
\hspace{1.5in}
         on(x,y,s)\land x \not= z.\\
poss(\movebtot(x,y),s) \liff  clear(x,s)\land on(x,y,s),\\
poss(\movettob(x,z),s) \liff  ontable(x,s)\land \\
\hspace{1.5in}     clear(x,s)\land clear(z,s).
\end{array}
$\\

Let $\mathbf{{\cal D}_{ss}}$ be a set of successor state axioms (SSA):\\
$
\begin{array}{c}
F(\bar{x}, do(a,\!s)) \leftrightarrow \gamma_F^{+}(\bar{x}, a,\! s) \lor \ 
	F(\bar{x}, s) \land \neg \gamma_F^{-}(\bar{x}, a,\! s),
\end{array}\\[0.5ex]
$
\noindent where $\bar{x}$ is a tuple of object arguments of the fluent $F$, and
each of the $\gamma_F$'s is a disjunction of uniform formulas
\ $[\exists\bar{z}]. a = A(\bar{u}) \land \phi(\bar{x},\bar{z},s)$,
where $A(\bar{u})$ is an action with a tuple $\bar{u}$ of object arguments, 
$\phi(\bar{x},\bar{z},s)$ is a context condition, and 
$\bar{z}\subseteq \bar{u}$ are optional object arguments; possibly $\bar{x}\subset \bar{u}$.
%
If $\bar{u}$ in an action function $A(\bar{u})$ does not include any $z$  variables, then there is no optional $\exists\bar{z}$  quantifier.
When $\bar{x}\subset \bar{u}$, that is the tuple of action arguments $\bar{u}$ contains all fluent arguments $\bar{x}$, and possibly contains $\bar{z}$, we say that the action $A(\bar{u})$ has a {\it local effect}. A BAT is called a {\it local-effect} BAT if all of its actions have only local 
effects. In a local-effect action theory, each action can change the fluent values only for objects explicitly named as arguments of the action. If (optional) conditions $\phi(\bar{x},\bar{z})$ do not mention $s$, and have no quantifiers, then SSA is called {\it context-free}. If there are no context conditions  $\phi$, then SSA is called {\it strictly context free} (SCF).
In BW, we consider fluents 
$clear(x, s)$, block $x$ has no blocks on top of it,
$on(x, y, s)$, block $x$ is on block $y$ in situation $s$, 
$ontable(x, s)$, block $x$ is on the table in $s$. 
The following SSAs are local-effect (with implicit outside $\forall x,\forall y,\forall a,\forall s$).
$
\begin{array}{l}
clear(x, do(a, s)) \leftrightarrow \\
     \hspace*{1mm}   \exists y,\!z(a\! =\! \movebtob(y,\! x,\! z))\llor
                    \exists y(a\!=\! \movebtot(y,\!x)) \lor\\ 
     \hspace*{10mm} 
     clear(x, s) \land \neg \exists y, z(a\! =\! \movebtob(y, z, x)) \land\\
          \hspace*{25mm} 
                 \neg \exists y(a\!=\! \movettob(y,x)),\\
on(x, y, do(a, s)) \leftrightarrow \\
      \hspace*{1mm}  \exists z(a\! =\! \movebtob(x, z, y)) \llor 
                      a\!=\! \movettob(x,y) \lor\\
        \hspace*{10mm}
    on(x, y, s) \land \neg\exists z(a\! =\! \movebtob(x, y, z)) \land\\
          \hspace*{25mm} 
                 a\!\not=\! \movebtot(x,y),\\
ontable(x, do(a,s)) \liff \exists y(a\!=\! \movebtot(x,y)) \lor\\
        \hspace*{10mm}
    ontable(x,s) \land \neg \exists y(a\!=\! \movettob(x,y)).\\
\end{array}\\
$

  ${\cal D}_{una}$ is a finite set of unique name axioms (UNA) for actions and
named objects.
For example,\\  
$
\begin{array}{c}
     \movebtob(x,y,z)\not= \movebtot(x',y'),\\
\movebtot(x,y)\!=\!\movebtob(x',y')\imply
            x\!=\!x'\land y\!=\!y'.\\
\end{array}
$

${\cal D}_{S_0}$ is a set of FO formulas whose only situation term is $S_0$. 
Specifies the initial values of the fluents. 
\commentout{
Also, it includes {\it domain closure for actions} such as 
$
\begin{array}{l}
\forall a. \ \exists x,y,z(a\eq \movebtob(x,y,z))\ \lor \\
            \hspace{10mm}   \exists x,y(a\eq \movebtot(x,y) )\, \lor\\
            \hspace{20mm} \exists x,y(a\eq \movettob(x,y) ).
\end{array}
$\\
}
\commentout{
In particular, it may include axioms for domain specific constraints 
(state axioms), e.g.,\\
$
\begin{array}{l}
\forall x\forall y( on(x,y,S_0)\imply \neg on(y,x,S_0) )\land \\
\forall x\forall y \forall z( on(y,x,S_0)\land on(z,x,S_0)\imply y\!=\!z )\land \\
\forall x\forall y\forall z( on(x,y,S_0)\land on(x,z,S_0)\imply y\!=\!z ).
\end{array}
$\\
Notice we did not include any state constraint (axioms uniform in $s$) into BAT,
e.g., 
$\forall x\forall y\forall z\forall s.\ on(y,x,s)\land on(z,x,s)\imply y\!=\!z$. 
As noted in \cite{Reiter2001}, they are entailed from the similar sentences 
about $S_0$ for any situation that includes only consecutively possible actions.
}

Finally, the foundational axioms $\Sigma$ are generalization of axioms for a single successor function (see Section 3.1 in \cite{Enderton}) since SC has a family of successor functions $do(\cdot,s)$: each situation may have multiple successors
\begin{description}
\item
\hspace*{-2mm}	$
	do(a_1,s_1)\!=\!do(a_2,s_2) \imply a_1\!=\! a_2\land s_1\!=\! s_2$
\item
\hspace*{-2mm} 	$\neg(s\sqsubset S_0)$
\vspace{-1mm}
\item
\hspace*{-2mm}	$
	s \sqsubset do(a,s') \liff s\sqsubseteq s'$,
where $s\sqsubseteq s' \isdef \ (s\sqsubset s' \lor s\!=\!s')$
\item	
\hspace*{-2mm}
$\forall P.\ \big(P(S_0)\land \forall a\forall s(P(s)\imply P(do(a,s)) )\, \big)
	\imply \forall s(P(s))$
\end{description}
The last second order (SO) axiom limits the sort \textit{situation} to the smallest set containing $S_0$ that is closed under the application of
$do$ to an action and a situation.
Therefore, the set of situations is really a tree: There are
no cycles, no merging. 
\commentout{
Since situations are finite sequences of actions, they can be implemented
as lists in PROLOG, e.g., $S_0$ is like [~], and $do(A,S)$ adds an action $A$
to the front of a list representing $S$, i.e. $[A | S]$.
Therefore, in PROLOG, situations satisfy $\Sigma$, and no SO reasoning is required to plan over situations. 
}

  It is often convenient to consider only executable (legal) 
situations: these are action histories in which it is actually possible 
to perform the actions one after the other.\\
$s \prec s' \isdef\, s\! \sqsubset s'\, \lland 
\forall a\forall s^* (\!s \sqsubset do(a,s^*\!) \sqsubseteq s'\imply poss(a,s^*))$\\
where $s \prec s'$ means that $s$ is an initial subsequence of $s'$ and all
intermediate actions are possible.
Subsequently, we use the following abbreviations: \ 
    $s \preceq s'\isdef (s \prec s') \lor s\!=\!s'$.\ 
Also, $executable(s)\isdef S_0 \leq s.$ \cite{Reiter2001} formulates 
\begin{theorem}\label{executability} 
$executable(s) \liff \big( (s\!=\!S_0) \lor$\\
\hspace*{10mm}
$\exists a\exists s'(s\!=\!do(a,s')\land poss(a,s') \land executable(s'))\, \big) $ 

\commentout{
$executable(do([\alpha_1,\cdots,\alpha_n],S_0)) \liff$\\
\hspace*{2mm}
$poss(\alpha_1,S_0)\land \bigwedge_{i=2}^{n}poss(\alpha_i, do([\alpha_1,\cdots,\alpha_{i-1}],S_0)).$
}
\end{theorem}
\begin{theorem}\label{relative-satisfiability}
\cite{PirriReiterACM1999}
A basic action theory 
${\cal D}\!=\!\Sigma \land {\cal D}_{ss} \land {\cal D}_{ap} \land {\cal D}_{una}\land {\cal D}_{S_0}$ 
is satisfiable \ iff \ ${\cal D}_{una}\land {\cal D}_{S_0}$ is 
satisfiable.
\end{theorem}
Thus, $\Sigma$ are {\it not} needed to check 
the satisfiability of $\bat$. 

The \textit{ Domain Closure Assumption} (DCA) for objects
\cite{ReiterJACM1980} means that the domain of interest is finite, the names of all objects in $\dsz$ 
are explicitly given as a finite set of constants $C_1, C_2,\ldots,C_K$, and for any object variable $x$ we have $\forall x (x\eq C_1 \lor x\eq C_2\lor \ldots\lor x\eq C_K)$. We do not include DCA in this paper.
 According to the \textit{Closed World Assumption} (CWA), an initial theory ${\cal D}_{S_0}$ is conjunction of ground fluents, and all fluents not mentioned in ${\cal D}_{S_0}$ 
are assumed by default to be false \cite{ReiterJACM1980}. 
According to an opposite, \textit{Open World Assumption} (OWA), $\dsz$ can be more general; e.g., it can be in a proper form. 

There are two main reasoning mechanisms in SC. 
One of them relies on the regression operator.
Another mechanism called progression 
is responsible for reasoning forward, where after each action $\alpha$ the initial theory $\dsz$ is updated to a new theory $\dsa$. In this paper, we focus on progression. 

Let $\alpha$ be a ground action, and let $S_\alpha$ denote the situation term $do(\alpha,S_0)$. A progression 
$\bat_{S_\alpha}$ of $\bat_{S_0}$ in response to $\alpha$ is a set of sentences uniform in $S_\alpha$ such that for all queries about the future of $S_\alpha$, $\bat$ is equivalent to 
$(\bat - \bat_{S_0}) \cup \bat_{S_\alpha}$.
We will not present the formal definition of progression, but we cite the following important property of progression.
It turns out that queries about $\sa$ can be answered from $\dsa$ alone, $\Sigma$ and other axioms from $\bat$ are \textit{not} needed:
\begin{theorem}\label{prog}
\cite{LinReiter1997}
Let $\bat_{S_\alpha}$ be a progression of the initial theory to $S_\alpha$. 
For any sentence $\phi$ uniform in $S_\alpha$,
$\bat \models \phi$ \ iff \ $\bat_{S_\alpha}\models \phi$. 
\end{theorem}
\vspace*{-1.5mm}
As explained in  \cite{LinReiter1997}, this ``theorem informs us that $\dsa$ is a strongest postcondition of the precondition $\dsz$ wrt the action $\alpha$".
\commentout{
In general, progression $\dsa$ is SO \cite{LinReiter1997}. 
However, in a special case in the next section, Theorem~\ref{WCFprogression} implies $\bat_{S_\alpha}$ is in FO.  Therefore, this property of progression is a key to the tractability of queries about $do(\alpha,S_0)$.
}

\section{Bounded Proper Planning}
In this section, we first formalize the task of bounded proper planning and then present two examples of it. 

In this paper, we consider a special form of a proper theory in the form of a finite set of ground literals, which we call a finite grounded proper (FGP) theory. Let $\KB$ be FGP. For a predicate $P$, we let $K\!P$ denote $\{\bar{d}\mid P(\bar{d}) \in \KB\}$, the set of tuples where $P$ is known to be true and $K\!\neg P$ for $\{\bar{d}\mid \neg P(\bar{d}) \in \KB\}$, the set of tuples where $P$ is known to be false.
Then the base case of $V$ is simplified for a FGP $\KB$ as follows: 

$V[\KB, P(\vec{d})]=
\left\{ \begin{array}{ll} 1 & \mbox{if $\bar{d} \in KP$}\\
0 & \mbox{if $\bar{d}\in K\!\neg P$}\\
\frac{1}{2}& \mbox{otherwise}
\end{array}
\right. $.

Recall that conjunctive queries are a common class of formulas for retrieving information from a database, e.g., see  
\cite{ChandraMerlinSTOC1977,AbiteboulHullVianu1995}.
In this paper, we extend them with safe disequalities.

\begin{definition} 
An extended conjunctive query (ECQ) is of the form $\exists \vec{x}\phi(\vec{x},\vec{y})$, where $\phi$ is a conjunction of positive literals and safe disequalities, that is, disequalities between variables or variables and constants such that each variable has to appear in at least one positive literal in $\phi$.
\end{definition}

A sufficient condition for a formula in negation normal form to be in
$\NF$ is that any two literals in the formula are conflict-free, that is, either they have the
same polarity, or they use different predicates, or they use different constants at some argument position. 
Therefore,
extended conjunctive queries are in $\NF$. 
Thus, $V$ is sound and complete for ECQ.

\commentout{
\begin{definition} A successor state axiom is strictly context-free (SCF) if 
$\gamma_F^+(\vec{x},a)$ and $\gamma_F^-(\vec{x},a)$ are disjunctions of formulas 
of the form $\exists \vec{z}[a=A(\vec{y})]$, where $A$ is an action function, 
$\vec{y}$ contains $\vec{x}$ and $\vec{z}=\vec{y}-\vec{x}$. 
\end{definition}
}


\cite{LiuLevesqueIJCAI2005} proposes a method for the progression of proper theories for local-effect and context-complete action theories. 
Note that a strictly context-free (SCF) action theory is both local-effect and context-complete. 
In this paper, we define a generalization of SCF action theories, called weakly context-free, and generalize their result to compute the progression of finite grounded proper theories efficiently. 

\begin{definition} A successor state axiom for fluent 
$F(\bar{x},\!s)$  is weakly context-free (WCF) if 
$\gamma_F^+(\bar{x},a)$ and $\gamma_F^-(\bar{x},a)$ are disjunctions of formulas of the form 
$\exists \bar{z}\big(a\!=\!A(\bar{u}) \wedge \phi(\bar{y},\bar{u})\big)$, where action 
arguments $\bar{u}$ contain $\bar{z}$, but arguments $\bar{y}$ are those fluent
arguments which are not in the set of action arguments, $\bar{y}\!=\!\bar{x}-\bar{u}$,
and $\phi(\bar{y},\bar{u})$ is the conjunction (over variables $y_i$ in $\bar{y}$) 
of equalities $y_i\!=\!g_i(\bar{u})$, where $g_i(\bar{u})$ is a term
built using a situation-independent (computable) function $g_i$, that is,
\( \phi(\bar{y},\bar{z})= \displaystyle{\bigwedge_{y_i\in \bar{y}} } y_i\!=\!g_i(\bar{u})  \). 
An action theory is WCF
if each SSA is either WCF or SCF.
\end{definition}
In other words, we assume that the language of the WCF basic action theory $\bat$ includes finitely many situation-independent (computable) functions that may occur only in the context
formulas $\phi(\bar{y},\bar{u})$ of the SSAs. 
For example, we can consider functions such as additions and multiplications. They have a standard interpretation and, as usual, 
a BAT $\bat$ does not include any axioms defining their properties; see, e.g. \cite{Reiter2001}. In automated planning and automated reasoning, there is a similar common approach known as semantic attachments  \cite{WeyhrauchAI1980}.

Thus, in a SCF axiom, the tuples for which the truth value of a fluent changes can be read from the action arguments. But for a WCF axiom, the tuples for which  the truth value of a fluent changes can be determined from the action arguments by using the given functions. Note that a WCF SSA is not necessarily local-effect, as each fluent argument in $\bar{y}$ is not mentioned in $A(\bar{u})$. Next, we show that progression of finite grounded proper theories can be efficiently computed for WCF action theories.

Using unique name axioms and functions $g_i(\bar{u})$, the instantiation of a WCF SSA on a ground action can be significantly 
simplified. Suppose that the SSA for fluent $F(\bar{x},s)$ is WCF. Let $\alpha\!=\!A(\bar{C})$ be a ground action, and let $\mu(a)=\exists \bar{z}[a\!=\!A'(\bar{u}) \wedge \phi(\bar{y},\bar{u})]$ be a disjunct of $\gamma_F^*(\vec{x},a)$, where $*$ is $+$ or $-$. If $A$ is different from $A'$, then $\mu(\alpha)$ is equivalent to false. Otherwise, $\mu(\alpha)$ is equivalent to $\exists \bar{z}[\bar{u}= \bar{C} \wedge \phi(\bar{y},\bar{u})]$. Since $\bar{u}$ contains $\bar{z}$ and $\bar{y}=\bar{x}-\bar{u}$, by using functions $g_i(\bar{u})$ with standard interpretation, $\mu(\alpha)$ is logically equivalent to conjunction of equalities $\bar{x}=\bar{d}$, where 
$\bar{d}$ contains constants from $\bar{C}$, 
for each argument in $\bar{u}$, and 
other constants computed from given functions $g_i(\bar{C})$,
for each $y_i\not\in \bar{u}$.
Thus, $\gamma_F^*(\vec{x},\alpha)$ is equivalent to a formula of the following form: 
\(\bar{x}\!=\!\bar{d_1} \vee \ldots\vee \bar{x}\!=\!\bar{d_n}\);
$n$ is the number of disjuncts in $\gamma_F^*(\vec{x},a)$.\\
We will use $\gamma_F^*(\alpha)$ to denote the set $\{\bar{d_i} \mid i=1,\ldots,n\}$.

In general, the progression $\dsa$ is in second-order logic. However, it is in first order for SCF action theories \cite{LinReiter1997}. 
Although WCF is an extension of SCF, the progression of WCF actions remains in first order, similarly to SCF actions, except that we use functions $g_i$ to provide fluent arguments as explained above. 

\begin{theorem}\label{WCFprogression} 
Let $\at$ be weakly context-free, $\KB$ be FGP, and $\alpha$ be a ground action. Then the progression of $\KB$ wrt $\alpha$, ${\cal P}(\KB, \alpha)$, remains FGP, and it can be computed 
as follows:
\begin{equation*}
\begin{split}
& K\!F := K\!F -\gamma_F^-(\alpha)\cup \gamma_F^+(\alpha), \\
& K\!\neg F := K\!\neg F - \gamma_F^+(\alpha)\cup \gamma_F^-(\alpha). 
\end{split}
\end{equation*}
\end{theorem}
\begin{proof}
Follows directly from Theorems 4 and 6 in \cite{LiuLevesqueIJCAI2005}. Note that the set $\gamma_F^+(\alpha)$ includes tuples for which fluent becomes true thanks to $\alpha$, $\gamma_F^-(\alpha)$ are tuples for which fluent becomes false due to $\alpha$. ${\cal P}(\KB, \alpha)$ is in FO.
\end{proof}

Informally speaking, if a ground fluent literal appears in ${\cal P}(\dsz, \alpha)$ and it has as arguments the new constants that are  not mentioned in $\dsz$, then one can say that they represent created objects, while regarding the constants that previously occurred in 
$\dsz$, but are no longer mentioned in progression, one can say that they represent objects destroyed by $\alpha$.

We now define the kind of BATs we consider in this paper.

\begin{definition}
We say a BAT is \textit{proper} if the following holds:
(1) The initial KB is a finite grounded proper theory; 
(2) all action preconditions axioms are quantifier-free extended conjunctive queries;
(3) all SSAs are weakly context free. 
\end{definition}


We now formalize our bounded proper planning (BPP) problem. We use $Length(\sigma)$ for the number of actions in situation $\sigma$, i.e., 
 $Length(do([\alpha_1,\cdots,\alpha_N],S_0) )\eq N$ and $Length(S_0)\eq 0$.

\begin{definition}\label{BPP}
A bounded proper planning (BPP) problem is a triple $P=(\bat, G(s), N)$, where $\bat$ is a proper BAT, $G(s)$ is a goal formula, which is an extended conjunctive query
uniform in $s$ and without other free variables, and $N\geq 0$ is a bound. A solution to $P$ is a ground $\sigma$ with length $\leq N$ s.t. 
\begin{equation}\label{planningReiter}
\bat \models_{\calE} executable(\sigma) \land G(\sigma).
\end{equation}
\end{definition}

Next, we present examples of bounded proper planning. 

For many years, British TV has run a popular show Countdown that includes the number round. 
\commentout{
Its rules are extremely simple. There are six initial  numbers drawn from a set of numbers that contains all numbers from 1 to 10 (small numbers) plus 25, 50, 75 and 100 (large numbers). With these six numbers, the contestants have to produce a target number randomly drawn between 101 and 999, or, if it is impossible, the closest number to the target number. Only the four standard arithmetic operations can be used. As soon as two numbers have been used to make a new one, they can’t be used again, but the new number calculated from those two can be used next. For our purposes, we modify this planning problem as follows.
}
We consider a version of Countdown that is modified for our purposes. The task is to produce the target integer number from a few given positive integers using only standard arithmetical operations. Once two numbers have been used in calculations, they are no longer available,  but the new number calculated from those two can be used next.
We allow any positive integer number as an initial, an intermediate, or a target number. For simplicity, consider only addition and multiplication.

We encode the modified Countdown problem using  fluents $available(c,s)$, which means that a counter $c$ is available in situation $s$, and $value(c,n,s)$ which means that a counter $c$ holds the number $n$ in $s$.  
The action
$add(c_1,v_1, c_2,v_2)$ adds the value $v_2$ of the counter $c_2$ to the value $v_1$ of the counter $c_1$, with the result stored in the counter $c_1$, overwriting the previous value stored there and making $c_2$ unavailable. 
The action $mult(c_1,v_1, c_2,v_2)$ represents the multiplication of values $v_1,v_2$ and the storage of the result in $c_1$. 
In the following preconditions, the right-hand sides are quantifier-free ECQ: 
\vspace*{-1.5mm}
\begin{equation*}
\begin{split}
	poss&(add(c_1,v_1, c_2,v_2), s) \leftrightarrow \\ 
    &available(c_1,s)\land  available(c_2,s)\land  c_1 \not= c_2 \land\\
	& value(c_1,v_1,s) \land value(c_2,v_2,s).
\end{split}
\end{equation*}
\vspace*{-2mm}
\begin{equation*}
\begin{split}
	poss&(mult(c_1,v_1, c_2,v_2), s) \leftrightarrow \\ 
    & available(c_1,s)\land  available(c_2,s)\land  c_1 \not= c_2 \land\\
	& value(c_1,v_1,s) \land value(c_2,v_2,s).
\end{split}
\end{equation*}

The successor state axioms for fluents are WCF:
\vspace*{-1mm}
\begin{equation*}
\begin{split}
	avai&lable(c, do(a,s)) \leftrightarrow \\
    & available(c,s) \land \neg\exists c',x,y (a\!=\!add(c',x,c,y))\land\\
	&\neg\exists c',x,y (a\!=\!mult(c',x,c,y)) .
\end{split}
\end{equation*}
\vspace*{-1mm}
\begin{equation*}
\begin{split}
	va&lue(c_1,v, do(a,s)) \leftrightarrow \\
	& \exists c_2,v_1,v_2 (a\!=\!add(c_1,v_1,c_2,v_2)\land v=v_1+v_2)\lor \\
	& \exists c_2,v_1,v_2 (a\!=\!mult(c_1,v_1,c_2,v_2)\land v=v_1\times v_2)\lor \\
	& value(c_1,v,s) \land
	\neg \exists c_2,v_2 (a\!=\!add(c_1,v,c_2,v_2))\land \\
	& \hspace*{2.0cm} \land
	\neg \exists c_2,v_2 (a\!=\!mult(c_1,v,c_2,v_2))
\end{split}
\end{equation*}

Consider several different initial theories. 
For clarity, instead of using constant names such  as $C_{54}$, 
we simply write $54$ to represent this integer number. 
Similarly, we write $1$ (or $2$) to represent the first (the second) counter in actions.\\

\textbf{Example 1}.  Let $\dsz$ be the following proper theory\\
$
\begin{array}{ll}
	\forall c.(c\!=\! 1\lor c\!=\! 2) \rightarrow available(c,S_0)\\
	\forall c,v.(c\!=\! 1\land v=4\lor c\!=\! 2\land v=5) \rightarrow value(c,v,S_0)\\
\end{array}
$\\

Note that $D_{S_0}$ does not say which counters are not available and if there are any other counters and what their values might be. This $\dsz$ is an FGP theory and has infinite models with countably many counters.  If the goal formula requires $\exists c(value(c,20,\sigma))$, then this goal can be reached by performing a single action $mult(1,4,2,5)$, since in every (infinite) model of $D_{S_0}$ this action is possible and leads to the goal state, regardless of the values in the other counters.

\textbf{Example 2}. Let $\dsz$ be the following proper theory\\
$
\begin{array}{ll}
	\forall c .(c\!=\! 1\lor c\!=\! 2) \rightarrow available(c,S_0)\\
	\forall c,v .(c\!=\! 1\land v\!=\!4\lor c\!=\! 2\land v\!=\!5) \rightarrow \neg value(c,v,S_0)\\
\end{array}
$\\

If the goal formula requires $\exists c(value(c,20,\sigma))$, then in the model $\cal M$ of $D_{S_0}$ where the first counter holds the number 5, and the second counter holds the number 4, this goal can be reached by doing a single action $mult(1,5,2,4)$. However, in another model $\cal M'$ where both counters have say number 10, this action is not even possible. In this example, no action is possible in all models of $\dsz$. In particular, actions operating with the numbers, say, in counters 7 and 8 can be possible in some models, but not in all models.

Note that in the above examples, new objects are created via addition and multiplication, 
and DCA does not apply.

\textbf{Example 3}. 
The chopping tree example discussed in the paper on iterative generalized planning \cite{HuLevesqueKR2010} provides connections to that area. We adapt the encoding from \cite{LinKR2008}, since it does not use sensing actions. Fluent $down(t,s)$ means that the tree $t$ is down, $size(t,n,s)$ for a tree $t$ means that $n$ is the number of cuts needed to bring the tree down. The action $chop(t,m)$ decreases the tree $t$ size from $m$ to $m-1$. The following are the action precondition axioms and SSAs:\\
$\begin{array}{l}
poss(chop(t,m), s) \liff size(t,m,s) \land m\not= 0\\
size(t,n, do(a, s)) \liff \exists m (a\!=\!chop(t,m) 
    \land n\!=\!m-1)\ \lor\\
\hspace{20mm} size(t,n,s)\land a\!\not=\!chop(t,n)\\
down(t,do(a, s)) \liff a=chop(t,1)\lor down(t,s)
\end{array}
$\\
In our example, if we do not make DCA for trees, then there are  infinitely many different infinite models of $\dsz$ where the trees have different thicknesses. (In terms of numerical planning, each infinite model would set infinitely many numerical variables to the values that the model prescribes.)  
Suppose that in $\dsz$ we have $\forall x,t(t\!=\!T_{127} \land x\!=\!4 \rightarrow size(t,x,S_0))$
and 
$\forall x,t(t\!=\!T_{1009} \land x\!=\!16 \rightarrow \neg size(t,x,S_0))$, 
i.e., the size of the specific tree  $T_{127}$ is 4, the size of the tree $T_{1009}$ is not 16, and there are no axioms about any other trees or their sizes.
Then, every model of  $\dsz$ includes the fact  that the initial size of $T_{127}$ is 4 and a combination of other facts. 
It is easy to see that the sequence of 4 chop actions
(call it $\sigma_4$) 
cuts the tree $T_{127}$ down and solves the planning problem with the goal formula $\exists t(down(t,\sigma_4))$. 
These actions are consecutively \textit{possible wrt all models}, 
including those infinite models that have countably many trees. 
However, if $\dsz$ is empty, i.e. it does not include any statements about fluent $size(t,x,S_0)$, or if it includes only the fact that the size of $T_{1009}$ is not 16, then 
this subtle modification has significant consequences. 
That is, there is no sequence 
of actions \textit{possible in all models} that leads to a ground situation $\sigma$, 
where the goal formula $\exists t(down(t,\sigma))$ holds. 
\commentout{
The previous work on iterative generalized planning discussed a program with a loop that in addition to a $chop$ action includes sensing actions, since this program can solve the problem wrt any model. In contrast, the BPP problem can be solved only if at each step there is at least one action possible wrt all models. 
}

\section{Solving Bounded Proper Planning}
Since proper BATs have different models that may include facts about infinitely many objects, it is not immediately obvious how to search starting from $\dsz$ for a plan that satisfies a goal formula wrt all models of the BAT $\bat$. We propose to search over ground situations, since they represent sequences of instantiated actions. To make this search finitely feasible, we introduce an upper bound (a positive integer) on the number of actions,  assuming that if an instance has a plan solving the problem, then its number of actions is less than a provided bound.  At each step of planning, we have to find all possible actions, and we have to make sure that there are only finitely many ground actions that are possible wrt all models of the BAT $\bat$. (We cannot choose an action that is possible with some models, but not with other models.) Thus, there might be at most finitely many ground situations that we can ever consider, and if a solution exists, then our search will find it. 
Therefore, the following Lemma~\ref{finiteBranching} is important, since it guarantees that there are at most finitely many successors of each situation node. Recall that a BAT mentions finitely many action functions $A(\bar{x})$.
\commentout{
, and each precondition 
$\Pi_A(\bar{x},s)$ is a conjunction of fluents and safe dis-equalities. 
}

\begin{lemma}\label{finiteBranching} Let $\bat$ be a proper BAT, $\sigma$ be a ground situation such that 
$\bat \emodels executable(\sigma)$. For any action function $A(\bar{x})$, there are only finitely many tuples 
$\vec{d}$ such that $\bat \emodels poss(A(\bar{d}),\sigma)$. 
\end{lemma}
\vspace{-1mm}
\begin{proof} 
Let ${\cal P}(\dsz,\sigma)$ be the progression of $\dsz$ wrt ground actions in $\sigma$: it is an FGP theory according to Theorem \ref{WCFprogression}. 
Let the precondition axiom for $A(\bar{x})$ be $poss(A(\bar{x}), s)\liff \Pi_A(\bar{x},s)$. 
By Theorem \ref{prog}, $\bat \emodels poss(A(\bar{d}),\sigma)$ iff ${\cal P}(\dsz,\sigma)
\emodels \Pi_A(\bar{d},\sigma)$. Since $\Pi_A$ is a quantifier-free ECQ, 
that is, a conjunction of positive literals and safe dis-equalities, ${\cal P}(\dsz,\sigma)$ is equivalent to a finite set of ground literals, and the procedure $V(\cdot,\cdot)$ is complete for ECQ queries, it retrieves only finitely many $\vec{d}$ such that ${\cal P}(\dsz,\sigma)
\emodels \Pi_A(\bar{d},\sigma)$.
\end{proof}

Note that this proof does not require the DCA for objects. Fluent $available(\bar{x},S)$ can be used to track objects destroyed by previous actions, and this fluent can occur in preconditions. Even if objects can be constructed or destroyed, at each planning step the set of named  objects remains finite. They are represented with constants mentioned in ${\cal P}(\dsz,\sigma)$.


Definition~\ref{BPP} and Equation~\ref{planningReiter} imply that the
planner has to search over executable sequences of actions. The state space remains  implicit, since situations serve as symbolic proxies to states that, from the perspective of semantics, are infinite models since our FGP theories include countable many constants. In terms of syntax, each situation uniquely corresponds to a finite number of fluent literals that comprise a state.
Whenever a sequence of $i$ ground actions results in a situation   $do([\alpha_1,\cdots,\alpha_i],S_0)$,
to find the next action, the planner must check among the actions
$A_1(\bar{x_1}),\ldots,A_k(\bar{x_k})$ for which of the values of their object 
arguments these actions are possible.
Since this computation is done at run-time, but not before search starts, our planner is lifted. Note that in a proper BAT, it is easy to check if a conjunctive $G(\sigma)$ holds for a ground $\sigma$.

The second key observation is that an efficient  planner needs control that helps select for each situation the most promising possible action to do. This control can be provided by an A$^*$ algorithm that is based on a domain-independent heuristic function inspired by \cite{HoffmannNebel2001}. 

\begin{algorithm}[ht]
\caption{$A^*$ search over situation tree to find a plan}\label{alg1}
\algorithmicrequire  $(\bat,G,N)$ - a bounded proper planning problem \\
\algorithmicrequire $H$ - Heuristic function $H(\bat,\!G,\!d,\!S_n,\!St)$\\
\algorithmicensure  $S$ that satisfies (\ref{planningReiter}) \Comment{Plan is the list of actions in $S$}

\begin{algorithmic}[1] 
    \Procedure{Plan}{$\bat,G,N,H,S$}
        \State $PriorityQueue \gets \varnothing$ \Comment{Initialize PQ}
        \State $S_{0}.Val \gets (N+1)$
        \State $PriorityQueue.insert(S_0, S_{0}.Val)$
        \State $Init \gets \text{InitialState}(\dsz)$ \Comment{Initialize state}
        \While{$ \textbf{not } PriorityQueue.empty()$}
        \State $S \gets PriorityQueue.remove()$   
        \State $Now \gets \text{\ Progress}(Init,S)$ \Comment{Current state}
            \If{\text{\ Satisfy($Now,G)\ $}}
                \State \textbf{return} $S$ \Comment{Found list of actions}
            \EndIf
            \State $Acts \gets \text{FindAllPossibleActions}(Now)$
            \If{$Acts == \varnothing$}
                         \State \textbf{continue} \Comment{No actions are possible in $S$}
            \EndIf
            \For{$A_{i} \in Acts$}
                \State $S_n \gets do(A_i,S)$ \Comment{$S_n$ is next situation}
                \State $St \gets \text{\ Progress}(Now,A_i)$ \Comment{Next state}
                    \If{$\ \text{Length}(S_n) \geq N$}
                         \State \textbf{continue} \Comment{$S_n$ exceeds upper bound}
                    \Else \ $d \gets N\!-\!\text{Length}(S_n)$ \Comment{$d$ is depth bound}
                     \EndIf
                     \Comment /*Each action has cost 1*/
                \State \hspace*{-1mm}$S_n.Val \gets \text{Length}(S_n)\! +\! H(\bat,\!G,\!d,\!S_n,\!St)$
                \State $PriorityQueue.insert(S_n,S_n\mathbf{.}Val)$
            \EndFor
        \EndWhile
        \State \textbf{return} $False$\Comment{No plan for bound $N$}
    \EndProcedure
\end{algorithmic}
\end{algorithm}

The main advantage of this design is that the frontier stored in a priority queue consists of situations and their $f$ values calculated as the sum of the length of the situation (cost $g$) and its heuristic estimate $h$, i.e. for each ground $\sigma$, $f(\sigma)\!=\!g(\sigma) + h(\sigma)$
\footnote{$f$-value is a term  from the area of heuristic search, see \cite{GeffnerBonetBook2013,GhallabNauTraverso2004}. 
There are plan costs $g(s)$, the number of actions in $s$, and there are heuristic estimates ($h$ values) of the number of actions remaining before the goal can be reached. The total priority of each search node (in our case it is situation
$s$) is estimated to be $f(s)\!=\!g(s)\!+\!h(s)$. A smaller total effort $f(s)$ indicates a more promising successor situation $s$.} 

 In Algorithm \ref{alg1}, 
 Line 7, the algorithm extracts the next most promising situation $S$ from the frontier. Then, on Line 8, it computes the progression $N\!ow$ of the initial state using the actions mentioned in $S$. In line 9, there is a check to see whether the goal formula $G$ is satisfied in the current state $N\!ow$. If it is, then $S$ is returned as a plan. If not, then on Line 12, the algorithm finds all actions that are possible from the current state using the precondition axioms. If there are no actions possible from $N\!ow$, then the algorithm proceeds to the next situation from the frontier. Otherwise, for each possible ground action $A_i$, it constructs the next situation $S_n\eq do(A_i,S)$, and if its length does not exceed the upper bound $N$, it computes the positive integer number $d$ on Line 21 as $N\!-\!Length(S_n)$. This bound $d$ is provided as an input to the heuristic function $H(\bat,G,d,S_n,St)$ that performs a  limited look-ahead up to depth $d$ from $St$ to evaluate situation $S_n$. In Line 24, $S_n$ and its $f$-value  $S_n.Val$ are inserted into the frontier, and then the search continues until the algorithm finds a plan or explores all situations
with at most $N$ actions. The \textbf{for}-loop, Lines 16-24, makes sure that all possible successors of $S$ are constructed, evaluated and inserted into the frontier. This is to guarantee the completeness of Algorithm~\ref{alg1}.

\begin{theorem} Algorithm \ref{alg1} is sound and complete for BPP. 
\end{theorem}
\begin{proof} 
By Lemma 1, there are only finitely many executable situations of length $\leq N$. 
The algorithm enumerates them in the nondecreasing order of the $f$ values. Thus, all of them will be enumerated and checked for satisfaction of the goal in Line 9. By Theorem \ref{prog}, $\bat \models_{\calE} G(do(\sigma,S_0))$ iff $\KB \models_{\calE} G$, where $\KB={\cal P}(\bat_{S_0},\sigma)$. 
By the soundness and completeness of $V$ for the ECQs, $V[\KB,G]=1$ iff $\KB \models_{\calE} G$. 
So, the algorithm is  sound and complete. Obviously, it terminates.
\end{proof}


The calculation of the heuristic function is done in two stages, with the usual delete relaxation, that is, the negative effects of actions are ignored, as in \cite{HoffmannNebel2001}.
We delegate the pseudocode of this algorithm to the Appendix, and the readers can find a detailed discussion in \cite{SoutchanskiYoungCEUR2023}. 
First, a planning graph is built from the current state, 
layer-by-layer until the goal is satisfied  (Line 4) 
or the state layer stops changing (Line 7). 
Supporting actions are then found for the goal literals, going backwards through the graph. We skip the details of how $Reachability(\bat,Goal,PG)$ is computed, since this is a minor improvement of the FF heuristic of \citeauthor{HoffmannNebel2001}. The interested readers can find these details in
\cite{SoutchanskiYoungCEUR2023}.
A minor adaptation of the proof in \citeauthor{HoffmannNebel2001} leads to 
\begin{theorem}
    Suppose that the state layer stops changing and the goal is
not satisfied. Then the original planning problem is not solvable.
\end{theorem}

\section{Implementation and Experiments}
We have developed a straightforward implementation of Algorithms~\ref{alg1} 
in PROLOG, following a logic programming approach from \cite{Reiter2001} and \cite{LevesqueThinking2012} with a few  important differences. Reiter and Levesque represented the initial theory as a collection of atomic statements for fluents that are true, but since PROLOG has a built-in CWA and DCA this is too restrictive for our purposes. Our implementation of proper BATs considers fluents as terms and collects initial proper KB literals into two lists of nodes, one node per fluent name, where each node includes a finite list of tuples. The first list of nodes represents those fluents that are initially known to be true, and the second list of nodes represents those fluents that are initially known as false. Both lists can be empty for some fluents, if the initial proper KB does not include any literals for those fluents. When we compute progression,  these two lists of nodes are updated accordingly to what the effects are of each action. Thus, we can work with infinite logical models by using two finite lists of tuples.

The existentially quantified conjunctive goal queries can be directly implemented as lists of atoms that may include variables, since in PROLOG a query variable is implicitly existentially quantified \cite{Lloyd2012}. When the program decides whether an action possible in a ground situation satisfies a goal formula, it evaluates the goal list of atoms wrt a progression of the initially theory. This does not require any search, since this is a linear-time algorithm wrt the list of goal atoms: either an atom is true wrt to progression, or it is false. 
The module that computes all possible actions at planning time is a potentially computationally intensive part of the program, as this task is related to the well-known problem of computing all answers to conjunctive queries  \cite{ChandraMerlinSTOC1977,AbiteboulHullVianu1995}. However, in our experiments, we considered only the benchmarks with short precondition axioms. 

Our planner, domain files and problem instances 
have been loaded, compiled and run with ECLiPSe Constraint Logic Programming System, Ver7.0 \#63 (x86\_64\_linux), released on April 24, 2022. We run it on a desktop computer equipped with an 11th Genuine Intel(R) Core(TM) i7-11700K CPU (3.60GHz) using only a single thread. 
PROLOG 
never exceeded 512M of the memory limit allocated by default. The program has a minimal memory footprint since we only stored short situations in our priority queue. 
\commentout{BEGIN_COMMENTOUT
} 
  We could not compare our program with alternative implementations, as we are not aware of other planners that can solve our benchmarks. 
\commentout{  
}

First, we tried to solve a few instances of the \textbf{Countdown} benchmark that was implemented according to the proper BAT specified above. The program could easily solve the instances with up to 3 counters. When we experimented with different instances, the planner took less than 1 s to find a solution with 2 actions. For instances that involved 4 counters, the planner could find a sequence of 3 actions within 5 min. 
When the bound $N$ was small, the planner could find a solution quickly, but otherwise it took a long time. For instances that had more than 4 counters, the planner could not find a solution within 10 min. 
To understand the issue, assume that there are $C$ counters and notice that the first counter may have up to $2(C-1)$ new values after performing all additions or multiplications once with the numbers in other counters. These new values can participate in operations with other new elements in other counters, and so on. The number of objects destroyed or created by actions grows rapidly, and this increases the complexity of finding a solution.

Second, we experimented with an \textbf{Infinite Blocks World}  (IBW) benchmark that is a minor variation of the usual BW domain. There are infinitely many blocks named with positive integers. In the initial situation, a few of them can be on the table, and a few other blocks can be on or above these blocks. In addition, there are a few other blocks that satisfy fluent $available(x,S_0)$. These blocks are not on the table, not on any other block, and they are not clear. However, there is an action $bring(x)$ that can take any available block, bring it to the table, and then this block becomes clear but no longer available. 
Subsequently, a block brought to the table can participate in any BW action, as usual. In addition, we introduced a new action $merge(b_1,b_2)$ that can take two different blocks that are $light$ in situation $s$, clear and on the table, convert one of them into a block that is $heavy$, while destroying another block, so that it becomes no longer available, not on  the table, and not clear. Both blocks participating in merging are no longer light after this action, but the usual BW actions have no effect on whether a block is light or heavy. In IBW, we uniformly set the upper bound $N$ to 100 for all instances we tried. All instances with 4 or 5 blocks available initially are easy when a goal formula requires a single tower; they take less than 1 sec. In the instance where 6 blocks are available, 
if the goal formula required heavy blocks as the base on the table, a plan with 11 actions was calculated in less than 1 min. The similar case, when blocks $\{1,2,3\}$ were initially in a tower on the table, while three more blocks were initially available, so that the fifth block must be a heavy foundation on the table of a new tower composed of the blocks $\{1,2,3,4,5\}$, a plan with 10 actions completed in a few seconds. Surprisingly, a few instances that included 7 blocks could not complete in 10 min. 

Finally, we have developed a new benchmark \textbf{Mixers} designed to illustrate planning with a dynamically changing domain. Mixers is a significant modification of the well-known Logistics benchmark. In Mixers, there are vehicles that can move between locations. In addition, there are ingredients and compounds of different types. They can be loaded into vehicles or unloaded under the usual preconditions for these actions in logistics. However, we added an additional condition that a vehicle must be empty to ensure that only one object can be loaded at a time. This is to guarantee that actions have local effects, since otherwise, in Logistics, the actions that move vehicles have global effects on several objects loaded into vehicles, but this is not allowed in this paper. 
Imagine any powders or liquids as ingredients and think about mixing them to produce a compound of a new type. Each type is a positive integer. Any ingredient available initially or intermediary compounds can be mixed according to the recipes given. After mixing, the starting ingredients are no longer available, but a compound produced becomes available for subsequent actions. 
\commentout{
The preconditions of action $mix(Mode, Loc, Obj1,Type1,Obj2,Type2)$ include that two participating objects must be at the same location, they have to be available, and there should be a recipe $recipe(Mode,Obj1,Type1,Obj2,Type2,Compound,New)$ specifying what kind of compound is produced from mixing, and what is its type, where mode specifies the recipe identifier.
}
Each mix action has a unique ID that corresponds to a recipe specifying the name and type of compound produced.
The new type is computed according to a specific numerical function. In our PROLOG implementation, we make sure that a new compound is named with a new string computed as the concatenation of the strings that name the starting ingredients. This is done to make sure that new domain elements are never mentioned initially and that the names of new elements are produced at planning time. Note that the objects can be created or destroyed by the mix actions, but without performing the actions, one cannot specify the objects in advance without extensive domain engineering. Therefore, this benchmark reflects well the purposes of planning with dynamically changing objects and without DCA. 
Some computational experiments conducted with instances demonstrated that Mixers  is challenging. We used small values of $N$. We easily solved the instances with 3 recipes, two locations, and 3 or 4 ingredients that required a few movement actions and 1 mix action. The plans with 4 or 5 actions completed in less than 1 sec. However, the plan that required doing two different mix actions at the same location was computed in about 10 min. Any more complex instances could not be solved within the 10min limit. This benchmark is challenging because the available objects change quickly when several mix actions are required.

  As a summary of our observations, we propose the hypothesis that planning with dynamically changing objects is a computationally intensive process, since the planner has to deal with the increasing number of combinatorial possibilities.
  We consider our planner as a proof-of-the-concept implementation and anticipate that significant work is required to design an efficient program that can plan without DCA.

\section{Discussion and Future Work}
The previous research on generalized planning focused on iterative plans, or programs that support the search for a plan that works at once for a set of classical planning problems. Our work considers sequential planning without the DCA. We solve the planning problem for goals that are conjunctive queries with $\exists$-quantifiers over object variables, but except for \cite{FrancesGeffnerIJCAI2016,FunkquistStahlbergGeffnerKR2024}, most previous work considered only conjunctions of ground fluents as goal formulas. Note that since we ground actions in preconditions, checking whether the resulting ground situation satisfies a goal query is easy in proper BATs.

Helmert \cite{Helmert2002} provides a comprehensive classification of the numerical
planning formalisms, demonstrates the cases where the planning problem is undecidable, 
and explores the reductions of more complex numerical planning formalisms to 
conceptually simpler cases. 
The numerical planning problems, where the range of values is finite, can be reduced to
classical planning with DCA; see, e.g. \cite{GiganteScalaIJCAI2023,BonassiPercassiScala2025}. 
In a general case, numerical planning goes beyond DCA. 
Our proposal is different, since we consider a BPP problem with \textit{incomplete} knowledge,
the models of progression in BPP can include the values of 
infinitely many numerical variables in contrast to numerical planning where 
each state is a finite set, and our actions can have parameters that range over
\textit{infinite domains}, while this is not allowed in PDDL \cite{HaslumLipovetzkyMagazzeniMuisePDDL2019}.

Note that in contrast to \cite{DeGiacomoLesperancePatriziAIJ2016}, we do not require that the number of objects where fluent can hold be bounded for all $s$. Informally, boundedness of the set of named objects that may ever be considered by our planner becomes the consequence of working with a proper BAT and 
imposing the upper bound on the number of actions.  

Lifted classical planning has been previously explored, 
e.g., see \cite{CorreaFrancesPommereningHelmertICAPS2021,Stahlberg2023}.
Our work is different since we focus on planning without  DCA. 

\cite{SoutchanskiYoungCEUR2023} proposed a lifted deductive planner  based on 
the situation calculus (SC), but their implementation  required both DCA and CWA. 
Their planner was competitive with Fast Downward \cite{HelmertJAIR2006,FD} in
terms of IPC scores based on the number of visited states and the length of the plan
(over classical planning benchmarks with a small number of objects).

\cite{CorreaDeGiacomoHelmertRubin2024} consider an extension of classical planning
(the domain of objects is \textit{finite}), with \textit{complete} knowledge, but 
their semantics is based on an unusual object assignment. 
Our BPP is more general, since we plan over infinite domains and consider incomplete knowledge.

The case of open-world planning is explored in \cite{BorgwardtEtAlKR2021,BorgwardtEtAlAAAI2022}.
They work with state constraints that are not explored in our approach. However, 
they restrict arities of fluents to use description logic ontologies, but our approach does not have restrictions on the arity. 

To our knowledge, there are no other heuristic planners that can solve
problems without the DCA given \textit{incomplete} initial theory. The conformant planners previously developed require DCA \cite{HoffmannBrafmanAIJ2006,PalaciosGeffnerJAIR2009,GrastienScalaAI2020}.
The planner in \cite{HoffmannBrafmanAIJ2006} was actually inspired by situation calculus, 
and it does search over sequences of actions, but it works only at a propositional level. 
\cite{Finzi00} does open-world planning, but they require DCA,
see details in \cite{Reiter2001}. 

Future work may consider the case where the identity of some objects is not known and, for this reason, a proper initial theory $\dsz$ may include $\exists$-quantifiers over objects; they can be replaced with Skolem constants. This case was discussed in \cite{DeGiacomoLesperanceLevesqueIJCAI2011}. 
However, there are no heuristic planners that support unknown individuals represented with Skolem constants.

\section{Appendix}
\begin{algorithm}[ht]
	\caption{GraphPlan heuristic with delete relaxation}\label{alg2}
	\algorithmicrequire $(\bat, G)$ - proper BAT $\bat$ and a goal formula $G$ \\
	\algorithmicrequire  $d\geq 1$ - Look-ahead bound for the heuristic algorithm \\
	\algorithmicrequire $S_n,L$ - The current situation and its length \\
	\algorithmicrequire $St$ - The current state \\
	\algorithmicensure  {\it Score} - A heuristic estimate for the given situation
	
	\begin{algorithmic}[1] 
		\Procedure{$H$}{$\bat,G,d,S_n,St$}
		\State $Depth \gets 0$ 
		\State $PG \gets \langle S_n,St\rangle$ \Comment{Initialize Planning Graph}
		\While{$\textbf{not} \: \textrm{Satisfy}(St,G)\ \textbf{and\ } Depth\leq d$}
		\State $\{ ActSet\} \gets \text{FindAllPossibleActions}(St)$
		\State $N\!ew\!Acts\! \gets \textrm{Select new possible actions from} \ ActSet$
         \If{$\ N\!ew\!Acts== \emptyset$}
         \Comment{Goal is unreachable}
                \State \textbf{return} $(L+d +1)$  \Comment{Penalty}
        \EndIf
		\State $St \gets \textrm{ProgressRelaxed}(St,N\!ew\!Acts)$ \\
		     \Comment{Add all new positive effects $N\!ew\!E\!f\!f\!s$ to the state}
		\State $N\!ext\!Layer \gets \langle N\!ew\!E\!f\!f\!s,N\!ew\!Acts,St \rangle$\\
		 \Comment{Record actions added, their effects, the current state}
		\State $PG.extend(N\!ext\!Layer)$
		\State $Depth \gets Depth+1$ 
		\EndWhile
         \Comment{If \textrm{Satisfy}(St,G)\ succeeds, $G$ is grounded}
        \State $Goal \gets $ \textrm{Convert grounded $G$ into a set of atoms}  
         \If{$Depth > d$} \textbf{return} $(L+d)$   \Comment{Penalty}
         \Else\ \ \textbf{return} $Reachability(\bat,Goal,PG)$
                     \EndIf
		\EndProcedure
	\end{algorithmic}	
\end{algorithm}

\begin{algorithm}[h]
	\caption{Reachability score for a set of goal literals}\label{alg3}
	\algorithmicrequire  $(\bat,G)$ - A proper BAT $\bat$ and a set $G$ of goal literals \\
	\algorithmicrequire $PG$ - A planning graph, initialized to $\langle S_n,St\rangle$ \\
	\algorithmicensure  $V$ - A heuristic estimate for achieving $G$
	
	\begin{algorithmic}[1]
		\Procedure{$Reachability$}{$\bat,G,PG$}
		\If{\ $PG==\langle S_n,St\rangle$}  \ 	\textbf{return} 0
		\Else \, $\langle Ef\!f\!s,\! Acts,\! St \rangle\! \gets PG.removeOuterLayer$
		\EndIf
		\State \hspace*{-3mm}
$CurrGoals\! \gets\! G \cap E\!f\!f\!s$ \, \Comment{The set of achieved goals}
		\State \hspace*{-1mm}
$NewGoals \gets \emptyset$      \Comment{To collect preconditions}
		\State \hspace*{-1mm}
$BestSupport \gets \emptyset$      \Comment{Easiest causes for CurrGoals}
		\For{$g \in CurrGoals$}
		\State \hspace*{-3mm}
		 $Relev\! \gets$ \{actions from $Acts$ with $g$ as add effect\}
		\For{\ $a \in Relev$\ }
		\State $a.Pre \gets $ \{the set of preconditions of $a$\}
		\State $a.Estimate\! \gets Reachability(\bat,\!Pre,\!PG)$
		\EndFor
		\State $Best\!Act \gets$ \textrm{ArgMin} \{$a.Estimate$ over $Relev$\}\\
		 \Comment{Find the action from $Relev$ with minimum estimate} 
		\State $NewGoals \gets NewGoals \cup Best\!Act.Pre$
		\State $BestSupport \gets BestSupport \cup Best\!Act$
		\EndFor
		\State $RemainGoals \gets G - CurrGoals$
		\State $NextGoals \gets RemainGoals \cup NewGoals$
		\State $C_1 \gets \textrm{Count}(BestSupport)$ \Comment{i.e. \# of best actions}
		\State $C_2 \gets Reachability(\bat,NextGoals,PG)$
		\State \textbf{return} $C_1 + C_2$
		\EndProcedure
	\end{algorithmic}
\end{algorithm}

\bibliographystyle{named}

\bibliography{AIxIA2023}

\end{document}